\documentclass[sigconf]{acmart}
\AtBeginDocument{%
  }

\setcopyright{none}
\acmYear{2025}
\acmConference[KDD-UMC '25]{Undergraduate and Master’s Consortium at KDD 2025}{August 03--07,
  2025}{Toronto, Canada}
\acmISBN{978-1-4503-XXXX-X/25/08}

\usepackage{cleveref}
\usepackage{subfig}
\usepackage{float}
\usepackage{placeins}

\begin{document}

\title{Learning to Plan via Supervised Contrastive Learning and Strategic Interpolation: A Chess Case Study}

\author{Andrew Hamara}
\email{andrew_hamara1@baylor.edu}
\orcid{0009-0002-5843-5827}
\affiliation{
  \institution{Baylor University}
  \city{Waco}
  \state{Texas}
  \country{USA}
}

\author{Greg Hamerly}
\email{greg_hamerly@baylor.edu}
\orcid{0000-0002-0360-1544}
\affiliation{
  \institution{Baylor University}
  \city{Waco}
  \state{Texas}
  \country{USA}
}

\author{Pablo Rivas}
\email{pablo_rivas@baylor.edu}
\orcid{0000-0002-8690-0987}
\affiliation{
  \institution{Baylor University}
  \city{Waco}
  \state{Texas}
  \country{USA}
}

\author{Andrew C. Freeman}
\email{andrew_freeman@baylor.edu}
\orcid{0000-0002-7927-8245}
\affiliation{
  \institution{Baylor University}
  \city{Waco}
  \state{Texas}
  \country{USA}
}

\renewcommand{\shortauthors}{Hamara et al.}

\begin{abstract}
  Modern chess engines achieve superhuman performance through deep tree search and regressive evaluation, while human players rely on intuition to select candidate moves followed by a shallow search to validate them. To model this intuition-driven planning process, we train a transformer encoder using supervised contrastive learning to embed board states into a latent space structured by positional evaluation. In this space, distance reflects evaluative similarity, and visualized trajectories display interpretable transitions between game states. We demonstrate that move selection can occur entirely within this embedding space by advancing toward favorable regions, without relying on deep search. Despite using only a 6-ply beam search, our model achieves an estimated Elo rating of 2593. Performance improves with both model size and embedding dimensionality, suggesting that latent planning may offer a viable alternative to traditional search. Although we focus on chess, the proposed embedding-based planning method can be generalized to other perfect-information games where state evaluations are learnable. All source code is available at \url{https://github.com/andrewhamara/SOLIS}.
\end{abstract}



\received{15 May 2025}

\maketitle

\section{Introduction}

The ability to plan and reason through complex decisions is a defining characteristic of intelligence. Chess has long served as a landmark of both human and artificial intelligence, as mastery of the game requires a combination of memory, strategic planning, and calculation. Its strict rules and combinatorial complexity make it an ideal domain for studying structured decision making in an environment where exhaustive search is impractical. State-of-the-art chess engines such as AlphaZero~\cite{alphazero} and Stockfish~\cite{stockfish} achieve superhuman performance by pairing neural network regressors with powerful tree search. Their mastery, broadly speaking, arises from the capacity to search deeply and efficiently through vast branches of possible continuations.

In contrast, expert human players seldom depend on deep search~\cite{giraffe}. Instead, they identify strong moves via intuition developed over years of analysis and validate their choices with shallow lookahead. As a result, while engines are invaluable for tasks such as opening preparation and post-game analysis, they fail to replicate the reasoning patterns of elite players~\cite{giraffe, maia, maia2}. This mismatch limits their utility as training tools and has prompted growing interest in engines that better reflect human opponents. Models such as Maia and Maia2~\cite{maia, maia2} pursue this goal by framing human alignment as a classification task, selecting moves that humans are most likely to play. Rather than emulating human decisions, we seek to explore whether a learned latent space can support a more efficient, human-like search process that reflects the selectivity of human intuition.

To this end, we introduce a chess engine built on a transformer encoder that replaces regression and deep search with a directional planning process. We train our model using supervised contrastive learning to embed board positions into a continuous latent space aligned with state-of-the-art engine evaluations. Within this space, distance reflects evaluative similarity, and linear interpolation enables interpretable trajectories between game states. Crucially, move selection occurs entirely within the embedding space by selecting actions that advance toward known favorable regions.

Despite using only a shallow search of six half-moves, our model achieves an estimated Elo rating of 2593. We show that performance scales with model size and embedding dimensionality, and that the learned space supports meaningful embedding arithmetic. These results suggest that contrastive training may offer a viable path toward structured representations that support efficient planning, without the need for deep search.

Although our experiments focus on chess, the underlying approach of learning evaluation-aligned embeddings and selecting actions by advancing through latent space is more general. It extends to other zero-sum, perfect-information games such as Go and Shogi. In these settings, where position evaluations are available or are learnable, directional planning provides an efficient alternative to deep search.

\section{Background}

\subsection{Chess Programming}

In 1928, von Neumann~\cite{v1928theorie} introduced game theory to chess, describing it as a two-player, zero-sum, perfect information game to which his proposed fixed-depth minimax algorithm could apply. Several decades later, Shannon~\cite{shannon_chess} adapted the minimax algorithm for chess, guided by static evaluation functions of game states. In parallel, Turing~\cite{turing_chess} introduced his chess-playing ``Machine,'' the most advanced engine of its time, though too complex for the hardware available to him.

In 1997, IBM successfully scaled chess computing with their Deep Blue engine to defeat the reigning world chess champion, Garry Kasparov, in a head-to-head match. Notably, it is estimated that Deep Blue evaluated roughly 200 million positions per second~\cite{deepblue1, deepblue2}, compared to Kasparov's 5~\cite{giraffe}.

Early versions of Stockfish~\cite{stockfish} followed a similar approach to Deep Blue: handcrafted evaluation functions paired with deep alpha-beta tree search. The methodology was compelling and achieved superhuman performance, though it became clear that the limiting factor was the combined chess mastery of those designing the static evaluation function.

The rating bottlenecks associated with static evaluation inspired a vein of research independent of human guidance. AlphaZero~\cite{alphazero} and its open-source counterpart Leela Chess Zero~\cite{leela} made a significant leap to engines that learned to evaluate positions entirely through self-play reinforcement learning, with no prior knowledge beyond the rules of chess. Similarly, modern versions of Stockfish adopted an efficiently updatable neural network evaluation module (NNUE)~\cite{nnue} to accelerate position evaluation while maintaining traditional alpha-beta search.

While these advances have produced engines that far surpass human players in strength, the underlying search algorithms remain largely unchanged. They still follow principles first proposed by von Neumann and Shannon and later integrated into Deep Blue, relying on deep alpha-beta search through millions of states~\cite{giraffe}. Yet it remains remarkable that human players, searching only a few continuations per second, display an understanding that engines achieve only by evaluating millions~\cite{giraffe}. In this work, we explore whether contrastive learning can enable a more efficient search process by treating planning as traversal through an embedding space.

\subsection{Contrastive Learning}
\label{sec:background}

Contrastive learning of representations encourages a model to embed similar inputs closer together and dissimilar inputs farther apart in a latent space~\cite{contrastive}. More concretely, given a batch $I$ containing an anchor input $x_i$, an encoder $f(\cdot)$ maps the anchor to a normalized embedding $z_i = f(x_i)$. Positive pairs $(z_i, z_j)$ correspond to semantically similar inputs, while all other pairs of samples in the batch are considered negatives.

A common contrastive objective is InfoNCE~\cite{infonce}, defined as:

\begin{displaymath}   
\ell_{i,j} = -\log \frac{\exp(\text{sim}(z_i, z_j) / \tau)}{\sum_{k \in A(i)} \exp(\text{sim}(z_i, z_k) / \tau)},
\end{displaymath}

where $A(i) = I \setminus \{i\}$ is the set of all samples in the batch excluding the anchor $i$, $\text{sim}(\cdot, \cdot)$ denotes a similarity metric, and $\tau$ is a scalar temperature parameter. In the case of self-supervised contrastive learning, the positive sample $z_j$ is an augmentation of the anchor $z_i$, and labels are not required~\cite{contrastive, supcon}.

When labels are available, the supervised contrastive loss (SupCon)~\cite{supcon} extends InfoNCE by allowing multiple positives per anchor. The SupCon loss is defined as:

\begin{displaymath}
\mathcal{L}_{\text{sup}} = \sum_{i \in I} \frac{-1}{|P(i)|} \sum_{p \in P(i)} \log \frac{\exp(\text{sim}(z_i, z_p)/\tau)}{\sum_{a \in A(i)} \exp(\text{sim}(z_i, z_a)/\tau)},
\end{displaymath}

where $P(i)$ denotes the set of positive examples corresponding to anchor $i$, and $A(i)$ is defined as above. Increasing the number of positives and negatives has been shown to improve representation quality~\cite{supcon, hamerly_aquatic}.

\section{Methods}

We now describe the components of our approach, including tokenization, model architecture, training procedure, and action selection during inference.

\subsection{Input Representation and Tokenization}

Each chess position is represented using its Forsyth-Edwards Notation (FEN) string, which compactly encodes the piece placement, side to move, castling rights, en passant target, half-move clock, and full-move counter. Following the tokenization scheme proposed in Ruoss et al.~\cite{searchless_chess}, we tokenize each FEN into a fixed-length sequence of 77 tokens by expanding run-length encodings.

\subsection{Encoder Architecture}

Our encoder is a multi-layer transformer~\cite{transformer} adapted for tokenized chess positions. We denote the number of transformer layers as $L$, the hidden dimension as $H$, the number of self-attention heads as $N$, and the output embedding dimension as $D$. We summarize the model configurations used in our experiments in Table~\ref{tab:model_details}.

\begin{table}[h]
\centering
\caption{Transformer model configurations used in our experiments.}
\label{tab:model_details}
\begin{tabular}{lcccccc}
\toprule
Model & $L$ & $H$ & $D$ & $N$ & MLP Size & Params \\
\midrule
Small & 6 & 512 & 512 & 16 & 512 & 8M \\
Base & 6 & 1024 & 1024 & 16 & 1024 & 41M \\
\bottomrule
\end{tabular}
\end{table}


Each input sequence is embedded into $H$-dimensional vectors through a learned token embedding matrix. A special classification token (CLS)~\cite{bert, vit} is prepended to the embedded sequence to aggregate information across the input. Because the input sequences are fixed length, we add a learned positional encoding to each token embedding~\cite{searchless_chess}. The resulting sequence is processed by $L$ stacked transformer encoder layers with GELU~\cite{gelu} activations and a dropout~\cite{dropout} rate of 0.1. The final hidden state corresponding to the CLS token is extracted, passed through a linear projection, and $\ell_2$ normalized.

\subsection{Supervised Contrastive Training}

We train our encoder using 5 million randomly sampled positions from the ChessBench dataset~\cite{searchless_chess}, each annotated with a Stockfish-evaluated win probability for the player to move. We normalize all evaluations to represent White's perspective, such that values near 1.0 indicate a decisive advantage for White and values near 0.0 indicate an advantage for Black.

To define positive samples for training, we set $\delta = 0.05$ as the evaluation margin for identifying similar positions. For each anchor, we precompute all positions whose win probabilities differ by less than $\delta$ and randomly sample five as positives during training. To build the batch-level mask, we compare the win probabilities of all samples and mark pairs as positive if their evaluation difference is below $\delta$.

We apply supervised contrastive learning (SupCon; see Section~\ref{sec:background}) using the constructed batch masks to define positive and negative pairs. Cosine similarity is used to compute pairwise scores, and we set $\tau = 0.07$~\cite{imagebind, clip} to control the sharpness of the softmax distribution. We train both models using stochastic gradient descent with a momentum parameter of 0.9~\cite{vit, contrastive, resnet} for 400,000 steps with a batch size of 128. Training is conducted on four NVIDIA L40S GPUs (48 GB memory each) with data parallelism~\cite{pytorch}.

\begin{figure}
\includegraphics[width=8 cm]{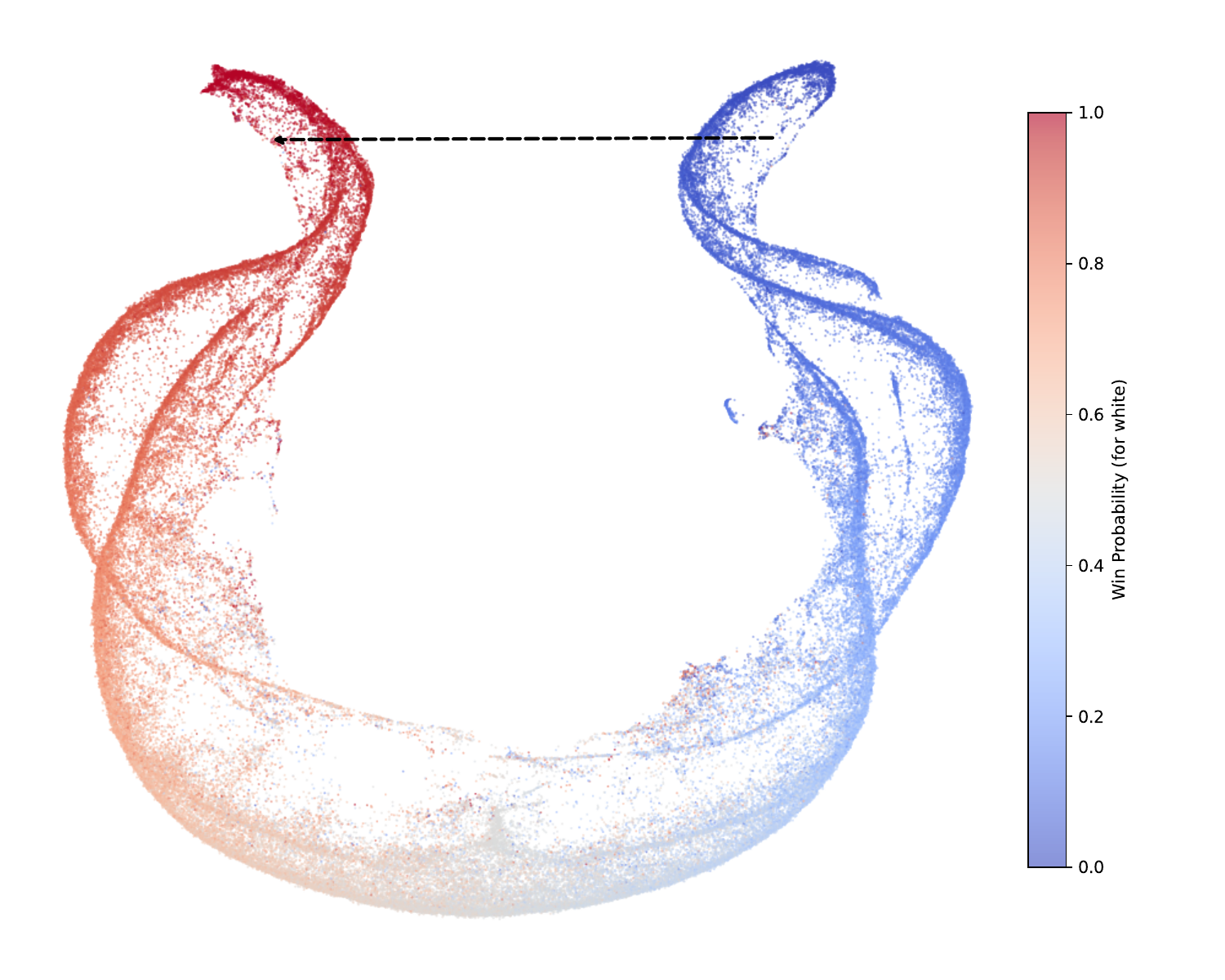}
\caption{UMAP projection of the learned embedding space of our Base model, colored by win probability (red = White favored, blue = Black favored). The dashed black arrow represents the advantage axis $\vec{a}$.}
\label{fig:embedding_space}
\end{figure}

We visualize the learned representation space of our Base model in Figure~\ref{fig:embedding_space} using UMAP~\cite{umap} for 2D projection. Each point represents an encoded chess position from a test dataset, colored by its Stockfish-evaluated win probability (for White). The black dashed arrow indicates the advantage axis $\vec{a}$, though its appearance in 2D does not reflect its true geometric trajectory.

\unskip

\subsection{Embedding-Guided Beam Search}

During inference, candidate moves are evaluated based on their alignment with an advantage axis. We first compute element-wise mean embeddings $\mu_\text{white}$ and $\mu_\text{black}$ over positions evaluated as $p=1.0$ (white has forced checkmate) and $p=0.0$ (black has forced checkmate), respectively, and define the advantage vector $\vec{a} = \mu_\text{white} - \mu_\text{black}$.

\begin{figure*}[h]
\centering
\includegraphics[width=\textwidth]{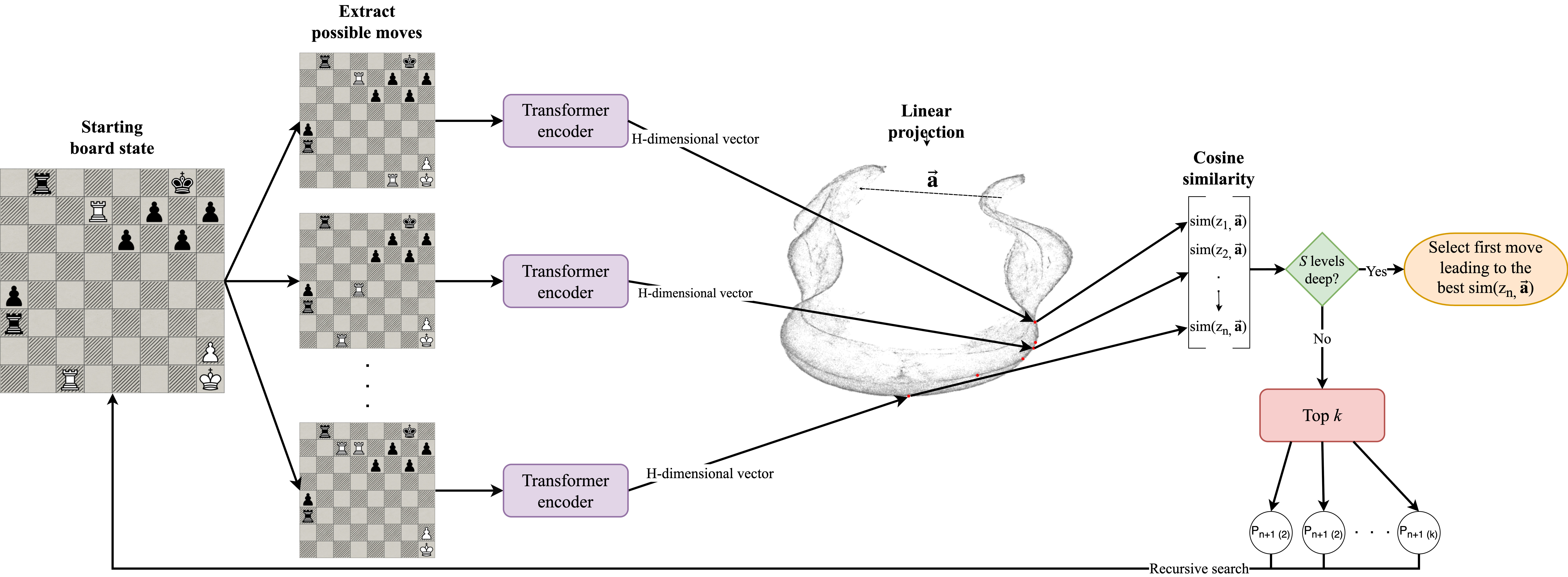}
\caption{System overview of the embedding-guided beam search. Candidate moves are embedded, scored via similarity to an advantage vector, and selected recursively based on top-$k$ alignment.}
\label{fig:system_overview}
\end{figure*}

As illustrated in Figure~\ref{fig:system_overview}, the search begins from a starting position (leftmost \emph{Starting board state}). All legal continuations are enumerated (\emph{Extract possible moves}) and tokenized. Each resulting child position is passed through our \emph{Transformer encoder} (purple blocks) in parallel, producing $H$-dimensional embeddings. Each candidate embedding $z'$ is linearly projected into the latent space and scored by its cosine similarity with the advantage vector $\vec{a}$: $\text{score}(z') = \cos(z', \vec{a})$ (\emph{Cosine similarity} block). The resulting scalar values reflect how promising each move is in terms of directional alignment with winning positions.

If the current search depth is equal to a predefined maximum $S$ (as checked in the green \emph{S levels deep?} decision node), the search terminates, and the initial move leading to the best final similarity score is selected (\emph{Select first move...}). Otherwise, the top-$k$ highest scoring candidate positions (\emph{Top-$k$} module) are selected and the process continues recursively, as shown by the loop labeled \emph{Recursive search}.

\section{Results}

\subsection{Experimental Setup}

Each model was evaluated against Stockfish 16 with a fixed per-move time limit of 50 ms~\cite{searchless_chess}. We configured our embedding beam search with a width ($k$) of 3 and ablated over search depths. At each depth, our model played at least 600 games against Stockfish configured with different Elo caps using its internal \texttt{UCI\_LimitStrength} setting. Caps were selected to ensure that each model configuration yielded at least one matchup with a positive win rate and one with a negative win rate. Elo ratings were computed using Bayesian logistic regression via the BayesElo~\cite{bayeselo} program with the default confidence parameter of 0.5. Summarized Elo estimates are presented in Table~\ref{tab:elo_summary}.

\begin{table}[h]
\centering
\caption{Estimated Elo ratings for our models by search depth and size.}
\label{tab:elo_summary}
\begin{tabular}{lcccccc}
\toprule
Model Size & Depth 2 & Depth 3 & Depth 4 & Depth 5 & Depth 6 \\
\midrule
Small & 2067 & 2282 & 2388 & 2487 & 2548 \\
Base  & 2115 & 2318 & 2433 & 2538 & 2593 \\
\bottomrule
\end{tabular}
\end{table}

\subsection{Performance Across Search Depths}

Our results show that Elo ratings increase consistently with search depth. The base model climbs from 2115 at depth 2 to 2593 at depth 6, roughly matching the strength of Stockfish configured to 2600 Elo. The small model follows a similar pattern with a consistent gap of 30–50 Elo relative to the base model.

Gains are nearly linear up to depth 5, but level off at depth 6. This tapering is likely due to the limitations of greedy beam search, which constructs plans incrementally. Because each step depends on the top-$k$ projections from the previous state, strong continuations outside this trajectory will be excluded. As a result, our model can miss advantageous lines that would only be considered via wider exploration.

\subsection{Qualitative Game Trajectories}

\begin{figure}
\centering
\subfloat[White steadily gains advantage.]
  {\includegraphics[width=4.0cm]{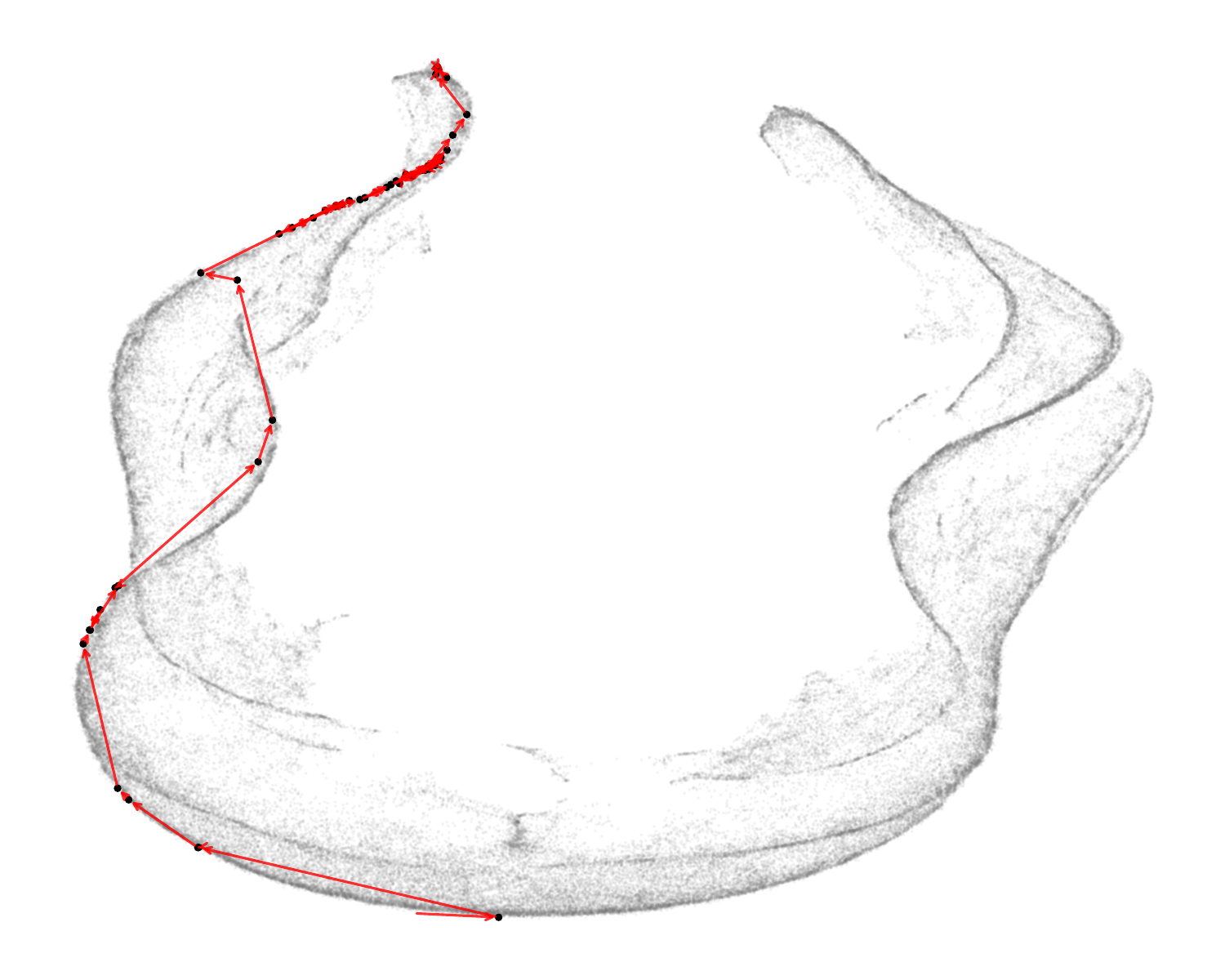}}
\hspace{0.3cm}
\subfloat[Black steadily gains advantage.]{\includegraphics[width=4.0cm]{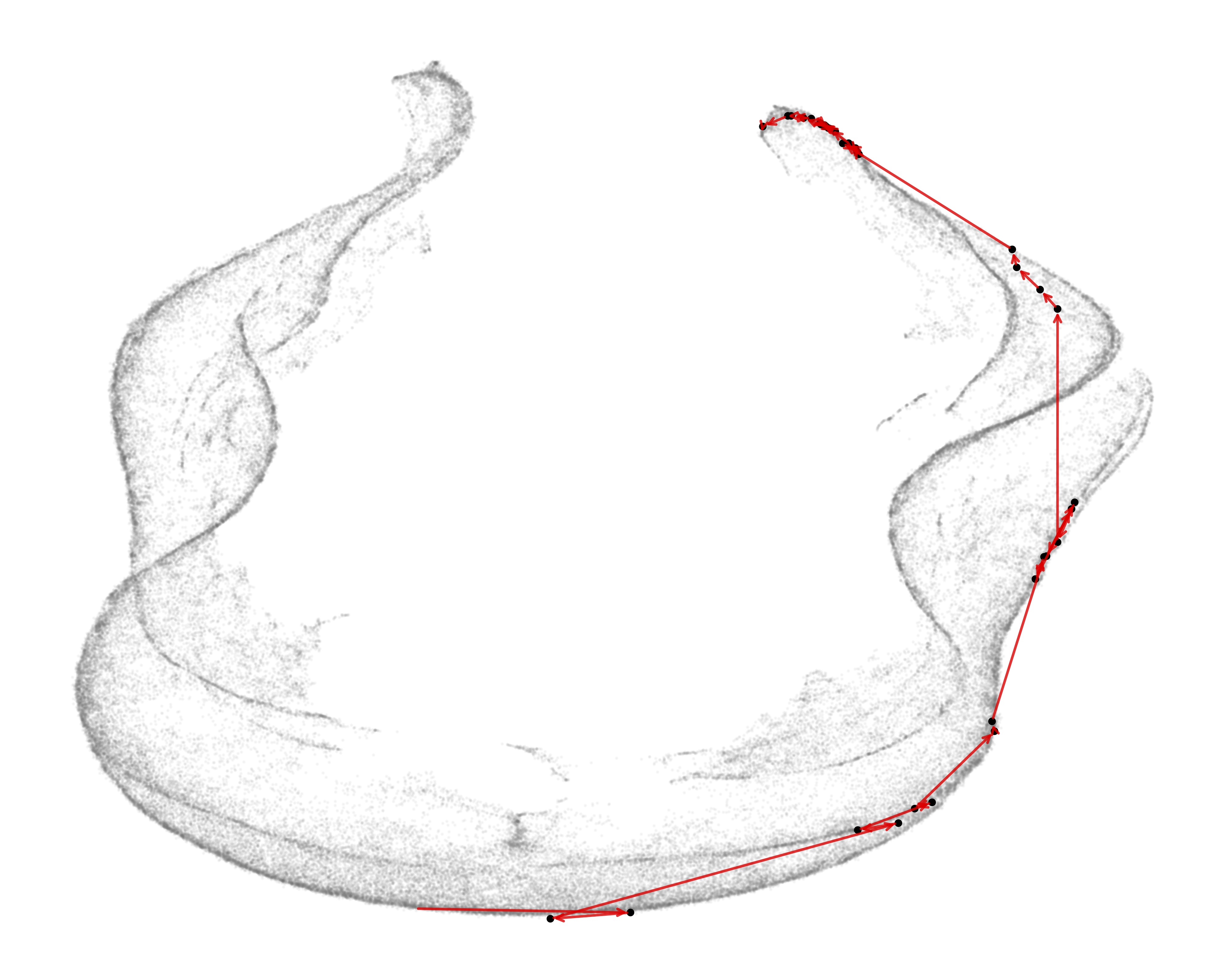}}
\hspace{0.3cm}
\subfloat[Game remains balanced throughout.]{\includegraphics[width=4.0cm]{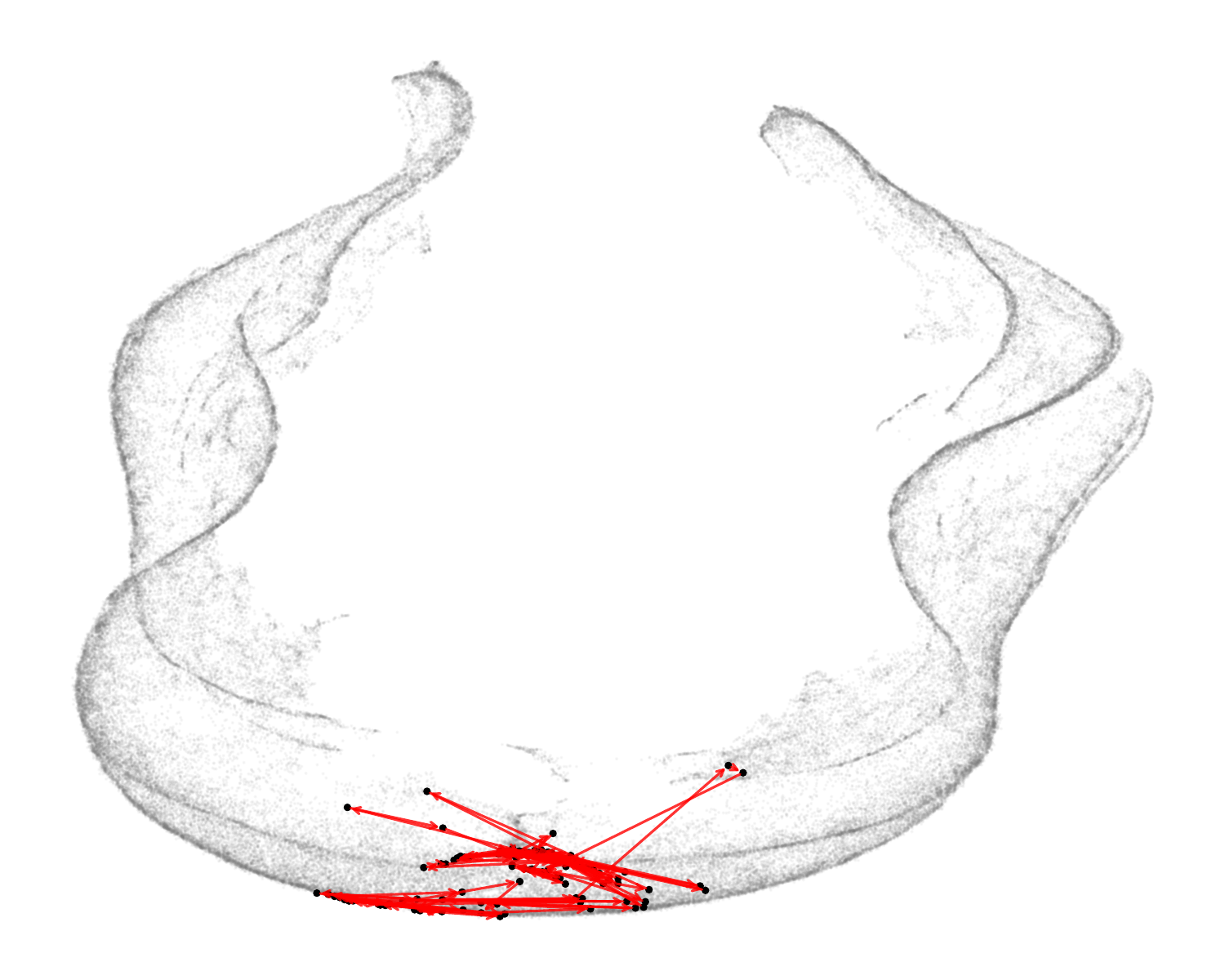}}
\caption{Latent trajectory visualizations of three games embedded in the shared representation space. Red arrows indicate the progression of positions as the game unfolds.}
\label{fig:game-trajectories}
\end{figure}

We visualize latent trajectories of real games to better understand the interpretability of the embedding space. Figure~\ref{fig:game-trajectories} shows three representative examples: a game won by White, a game that remains balanced throughout, and a game won by Black. Each trajectory is plotted over the same 2D projection from a sample dataset. Positions are embedded independently and connected with arrows to indicate move progression. Games that are decisively won by one player trace smooth paths through the space, while closely contested games fluctuate around the center.

\section{Discussion}

We have presented a chess engine that selects moves by advancing through a learned embedding space without relying on deep search. Despite using only a 6-ply search, the system achieves an Elo rating of 2593. Visualizations of real games further show that latent trajectories follow smooth evaluative trends.

While these results are promising, the system has notable limitations. Firstly, the greedy beam search cannot revise early commitments, restricting its ability to recover from errors. Broader planning strategies, such as policy networks, non-greedy beam exploration, and search memoization may improve robustness. Secondly, the current training pipeline assumes a fixed evaluation target derived from Stockfish, which may not reflect human intuition. Future work could explore reinforcement learning fine-tuning, larger models, and a broader sampling of positives to further shape the structure of the embedding space.

Overall, these results suggest that latent-space reasoning may offer a viable alternative to conventional search. We hope this work contributes to a broader rethinking of planning in games and decision-making domains, not as brute-force optimization but as strategic interpolation through learned representations.

\bibliographystyle{ACM-Reference-Format}
\bibliography{references}

\end{document}